%% file: main.tex
\documentclass{article}
\usepackage[final]{corl_2025} 
\usepackage{amssymb}
\usepackage{graphicx}
\usepackage{amsmath} 
\usepackage{natbib}  
\usepackage{hyperref} 
\usepackage{cleveref}
\usepackage{array}
\usepackage{booktabs}
\usepackage{wrapfig}

\usepackage{multirow}
\usepackage{listings} 
\usepackage{xcolor}   
\usepackage{subcaption} 
\usepackage{url}
\usepackage{float} 

\newcommand{\method}{SPF}
\newcommand{\methodfull}{See, Point, Fly}

\title{See, Point, Fly: A Learning-Free VLM Framework for Universal Unmanned Aerial Navigation}

\vspace{-5pt}
\author{
\textbf{Chih Yao Hu$^{2*}$ \quad}
\textbf{Yang-Sen Lin$^{1}\thanks{The first two authors contributed equally}$ \quad}
\textbf{Yuna Lee$^{1}$ \quad}
\textbf{Chih-Hai Su$^{1}$ \quad} 
\textbf{Jie-Ying Lee$^{1}$ \quad} \\
\textbf{Shr-Ruei Tsai$^{1}$ \quad}
\textbf{Chin-Yang Lin$^{1}$ \quad} 
\textbf{Kuan-Wen Chen$^{1}$ \quad}
\textbf{Tsung-Wei Ke$^{2}$ \quad}
\textbf{Yu-Lun Liu$^{1}$ \quad} \\
\vspace{0.1cm}
$^{1}$ National Yang Ming Chiao Tung University, $^{2}$ National Taiwan University \\
\\ \url{https://spf-web.pages.dev} \\
}

\lstdefinestyle{custompy}{
  language=Python,
  basicstyle=\ttfamily\small,
  keywordstyle=\color{blue},
  stringstyle=\color{red!60!black},
  commentstyle=\color{green!50!black},
  morecomment=[l][\color{magenta}]{\#},
  breaklines=true,
  showstringspaces=false,
  numbers=left, 
  numberstyle=\tiny\color{gray}, 
  frame=tb, 
  captionpos=b, 
}
\begin{document}
\maketitle



\input{1_abstract}

\keywords{Vision-Language Models, Zero-shot UAV Navigation, 2D-to-3D Waypoint Prompting}


\input{2_introduction}
\input{3_related_works}

\input{4_method}

\input{5_experiments}
\input{6_conclusion}







\clearpage

\textbf{Limitations.}
Despite promising results, our system has limitations. VLM inaccuracies (hallucinations and misinterpretations) can occur, and grounding precision may decrease for small or distant targets. The adaptive step heuristic provides implicit depth but can be imprecise. Performance can be sensitive to prompt phrasing. Reactivity to highly dynamic obstacles is limited by the VLM inference latency ($\approx$1-3s). Finally, VLM-generated search patterns are not guaranteed to be optimal. These limitations highlight avenues for future work, including improving perception robustness, improving grounding mechanisms, reducing system latency for better reactivity, exploring VLM fine-tuning, and developing more sophisticated exploration strategies.

\acknowledgments{This research was funded by the National Science and Technology Council, Taiwan, under Grants NSTC 113-2628-E-A49-023- and 111-2628-E-A49-018-MY4. The authors are grateful to Google, NVIDIA, and MediaTek Inc. for their generous donations. Yu-Lun Liu acknowledges the Yushan Young Fellow Program by the MOE in Taiwan.}


\bibliography{main}

\newpage
\clearpage

\appendix
\input{appendix}

\end{document}

%% file: 1_abstract.tex
\begin{abstract}
We present \methodfull{} (\method{}), a training-free aerial vision-and-language navigation (AVLN) framework built atop vision-language models (VLMs).  \method{} is capable of navigating to any goal based on any type of free-form instructions in any kind of environment.  In contrast to existing VLM-based approaches that treat action prediction as a text generation task, our key insight is to consider action prediction for AVLN as a 2D spatial grounding task.  \method{} harnesses VLMs to decompose vague language instructions into iterative annotation of 2D waypoints on the input image.  Along with the predicted traveling distance, \method{} transforms predicted 2D waypoints into 3D displacement vectors as action commands for UAVs.  Moreover, \method{} also adaptively adjusts the traveling distance to facilitate more efficient navigation.  Notably, \method{} performs navigation in a closed-loop control manner, enabling UAVs to follow dynamic targets in dynamic environments.  \method{} sets a new state of the art in DRL simulation benchmark, outperforming the previous best method by an absolute margin of 63\%.  In extensive real-world evaluations, \method{} outperforms strong baselines by a large margin.  We also conduct comprehensive ablation studies to highlight the effectiveness of our design choice.  Lastly, \method{} shows remarkable generalization to different VLMs.
Project page: \url{https://spf-web.pages.dev}

\end{abstract}

%% file: 2_introduction.tex
\vspace{-1mm}
\section{Introduction}
\vspace{-1mm}

The rapid development of unmanned aerial vehicles (UAVs) has revolutionized applications from environmental monitoring to security patrol. However, autonomous UAV navigation remains challenging due to requirements for strong visual reasoning in unstructured environments, language understanding for user instructions, and high-level task planning with low-level action control~\cite{chang2023review}. These autonomous UAV navigation tasks are often framed as aerial vision-and-language (AVLN) tasks~\cite{liu2023aerialvln,lee2024citynav}.

The autonomous UAV navigation tasks are commonly framed as aerial vision-and-language (AVLN) tasks~\cite{liu2023aerialvln,lee2024citynav}.  Conventional methods primalily adopt end-to-end policy learning frameworks which consist of a text and vision encoder that maps language instructions and visual observations into latent representations, followed by a policy head that converts these representations into UAV actions~\cite{blukis2018following,misra2018mapping,chen2023vision,liu2024navagent,fan2022aerial,wang2024towards,lykov2025cognitivedrone}.  The entire models are trained on a curated set of expert demonstrations~\cite{giusti2015machine,smolyanskiy2017toward,loquercio2018dronet,bozcan2020air}.  However, due to the limited scale and diversity of the training data, these methods fail to generalize to unseen environments or task instructions.  In contrast, recent works explore a training-free direction that directly converts Vison Large Language Models (VLM) into AVLN policies~\cite{chen2023typefly,gao2024aerial,xu2025geonav,sautenkov2025uav}.  As VLMs are trained on large-scale internet data, these models have demonstrated not only rich common-sense knowledge of the world, strong capabilities in visual/language understanding, reasoning and planning, but also, strong generalization to novel environments and tasks~\cite{achiam2023gpt,team2025gemini,yang2024qwen2}.

How to repurpose VLMs that generate texts into embodied agents that generate physical actions has attracted increasing interest in robotics~\cite{liang2023code,singh2023progprompt,huang2023voxposer,lin2023text2motion}, while the research direction is still underexplored in AVLN.  Existing VLM-based approaches to AVLN build atop a direct solution, that considers action prediction simply as a text-generation task.  VLMs are prompoted to output either continuous actions~\cite{gao2024aerial} or pre-defined skills~\cite{chen2023typefly,xu2025geonav,sautenkov2025uav}, in terms of texts.  Despite the simplicity of these methods, they have two obvious problems: (1) embodied agents need to execute fine-grained actions, while texts are not suitable to represent high-precision floating numbers; (2) these VLMs have not been trained on aerial navigation data to predict 3D actions for navigation. In contrast, our key insight is to consider action prediction for AVLN as a 2D spatial grounding task.  Instead of predicting 3D actions directly, we harnesses VLMs to annotate 2D waypoints~\cite{gu2023rt,liu2024moka,nasiriany2024pivot,hwang2024emma} on the image, which do not require any domain knowledge of AVLN but general spatial understanding~\cite{chen2024spatialvlm,cheng2024spatialrgpt}.  As these 2D waypoints are grounded in the visual scene, they inherently contain precise action information.  These 2D waypoints can then be transformed into 3D actions using the camera information.

Notably, we do not introduce the concept of predicting 2D waypoints for action selection—similar ideas have been explored in both robot manipulation and navigation~\cite{nasiriany2024pivot,gu2023rt,liu2024moka,team2025gemini}. For example, RT-Trajectory~\cite{gu2023rt} leverages VLMs to directly label 2D waypoints on the image, which are then used by a separately trained policy network to predict corresponding actions. PIVOT~\cite{nasiriany2024pivot}, in contrast, samples multiple candidate actions as 2D waypoints and employs a VLM to select the most appropriate one for execution. In this work, we build on this general idea and adapt it to the AVLN setting. Our method requires no additional neural network training, yet it significantly outperforms PIVOT, which is also a training-free approach.

\begin{figure}[t]
    \centering
    \vspace{-10mm}
    \includegraphics[width=1.0\linewidth]{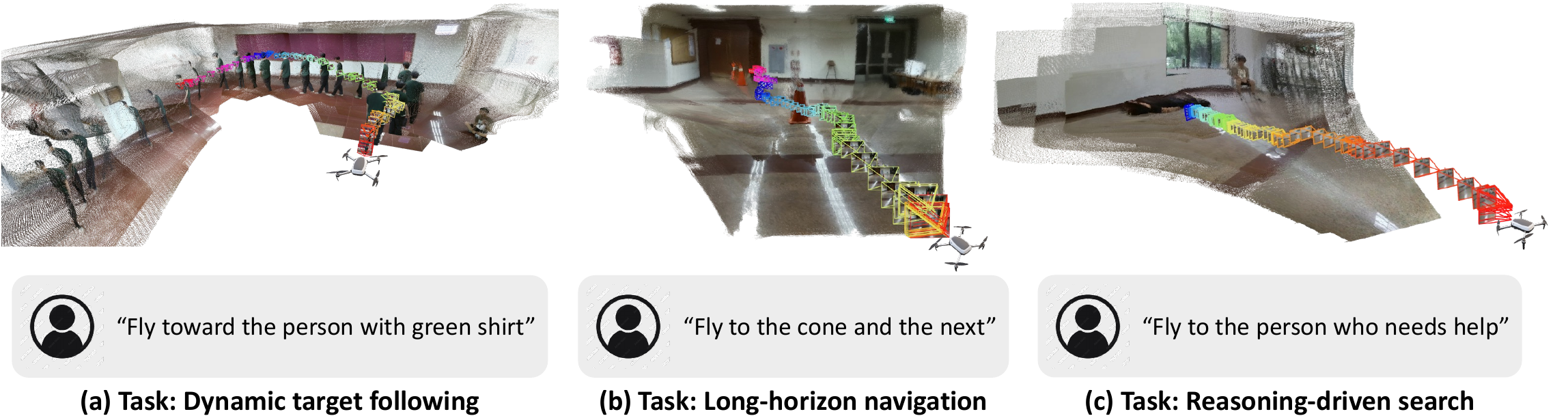}
    \vspace{-4mm}
    \caption{\textbf{Zero-shot language-guided UAV control.} (a) The UAV continually replans to keep pace with a moving person. (b) The UAV chains multiple goals across the hall. (c) The UAV locates the person on the ground and navigates around obstacles. Coloured 3D boxes mark successive camera viewpoints, revealing the UAV’s full flight trajectory over the reconstructed point cloud. All waypoints are generated directly by the vision-language model, with \emph{no} task-specific training.}
    \label{fig:teaser}
    \vspace{-4mm}
\end{figure}

We introduce \methodfull{} (\method{}), a novel VLM-based AVLN framework that navigates to any goal based on any free-form instructions in any environment.  At the core of our method is a VLM~\cite{team2025gemini} that conditions on the current scene and language instructions, and outputs the 2D waypoints in terms of pixel locations.  These 2D waypoints are unprojected into unit-length 3D positions based on the camera parameters.  These 3D positions denote the relative 3D actions to the current UAV location.  To enhance the navigation speed, we propose an adaptive controller module that adjust the scale of the actions based on the distance between the UAV and the target.  Since our method naturally enables closed-loop control of the UAV, as shown in Fig.~\ref{fig:teaser}, UAVs are capable of following dynamic targets.  Moreover, building atop VLMs, our method can easily tackle long-horizon and even ambiguous task instructions in a zero-shot manner.

We test \method{} on a simulation and a real-world benchmark.  Our method outperforms prior state-of-the-art, TypeFly~\cite{chen2023typefly} by a large margin.  We show that our method works well across a wide range of tasks, including long-horizon, abstract, and dynamic navigation tasks.  We also conduct an extensive ablation study to validate the effectiveness of each design choice.

In summary, our contributions are: (1) We propose a state-of-the-art AVLN framework that generalizes to novel scenes and free-form instructions; (2) We set a new state-of-the-art in the DRL simulator~\cite{DRLSim2024} simulation benchmark, outperforming prior SOTAs with a margin of $63\%$ in success rate; (3) We set a new state of the art in the real-world benchmark, outperforming prior SOTAs with a margin of $82\%$ in success rate.

%% file: 3_related_works.tex
\vspace{-1mm}
\section{Related Work}
\label{sec:Related work}
\vspace{-1mm}

\noindent \textbf{End-to-end policy learning in UAV navigation.}
The goal of policy learning is to train a model that outputs control actions for UAVs. Policy learning for UAV navigation can be broadly categorized into imitation learning (IL)~\cite{pomerleau1988alvinn} and reinforcement learning (RL)~\cite{sutton1998reinforcement}. The objective of RL is to maximize cumulative rewards through interaction with the environment. These methods have achieved strong performance in drone racing~\cite{kaufmann2023champion,song2023reaching,ferede2024end}, collision avoidance~\cite{kang2019generalization} and optimal quadrotor control~\cite{kaufmann2020deep,molchanov2019sim,hwangbo2017control,o2022neural,shi2019neural}. Recent work has also explored NeRF-based environments for validating autonomous navigation policies~\cite{shen2024driveenv}, providing realistic simulation environments for training and testing. However, RL often struggles with tasks involving long temporal horizons and sparse reward signals, and have shown limited success in navigation tasks.

On the other hand, the objective of IL is to maximize the likelihood of the actions from expert demonstrations~\cite{giusti2015machine,smolyanskiy2017toward,loquercio2018dronet,bozcan2020air}. Prior works focus on exploring effective policy architectures for navigation. GSMN~\cite{blukis2018following} proposes to construct intermediate map representations inside the policy, to facilitate action predictions. CIFF~\cite{misra2018mapping} utilizes a mask generator to annotate the goal location on the image, followed by recurrent neural network to predict the corresponding UAV actions. LLMIR~\cite{chen2023vision} and AVDN~\cite{fan2022aerial} instead build policies based on conditional transformers. Recent advances in robotic control have also demonstrated the effectiveness of diffusion-based methods for precise manipulation tasks~\cite{guo2024precise}, suggesting potential applications in UAV control. Notably, due to the limited capacity of language encoders inside these methods, they are incapable of handling free-form instructions in recent AVLN benchmarks~\cite{liu2023aerialvln,lee2024citynav,wang2024towards,gao2025openfly}. To enhance language understanding, recent works propose to fine-tune large language models as navigation policies~\cite{liu2024navagent,lykov2025cognitivedrone}.

While these end-to-end learning frameworks show good evaluation performance in similar settings as training data, due to the limited scale and diversity of the training data, these methods fail to generalize to unseen environments or task instructions. We instead explore a training-free alternative that deploy VLMs for AVLN in a zero-shot manner.

\textbf{Vison language models for training-free UAV Navigation.}
Converting VLMs, originally designed for text generation, into embodied agents that output action controls has drawn increasing attention.  A direct solution is to prompt VLMs to generate UAV actions in textual forms.  For instance, ~\cite{gao2024aerial} proposes to construct semantic map representation that localizes task-related objects in the bird's-eye aerial map, with VLMs.  Prompted with the map representations, VLMs output the corresponding 2D actions to reach the target on the map.  In stead of outputting continuous actions, TypeFly~\cite{chen2023typefly}, UAV-VLA~\cite{xu2025geonav}, Flex~\cite{chahine2024flex} and GeoNav~\cite{xu2025geonav} prompt VLMs to generate discrete actions, selected from a predefined set of navigation skills.  While both paradigms simplify the interface between language models and control systems, they restrict the UAV's action space, often leading to suboptimal motion trajectories and reduced control precision.  In stark contrast, our \method{} considers action prediction as a 2D spatial grounding task.  We utilize recent VLMs'~\cite{team2025gemini} strong capabilities in affordance annotation, prompting VLMs to label 2D waypoints~\cite{gu2023rt,liu2024moka,nasiriany2024pivot,hwang2024emma} on the image.  Transforming these 2D points into 3D actions with the camera information results in more effective UAV control.

%% file: 4_method.tex
\vspace{-1mm}
\section{Method}
\label{sec:method}
\vspace{-1mm}

\begin{figure}[t]
    \centering
    \vspace{-10mm}
    \includegraphics[width=1.0\linewidth]{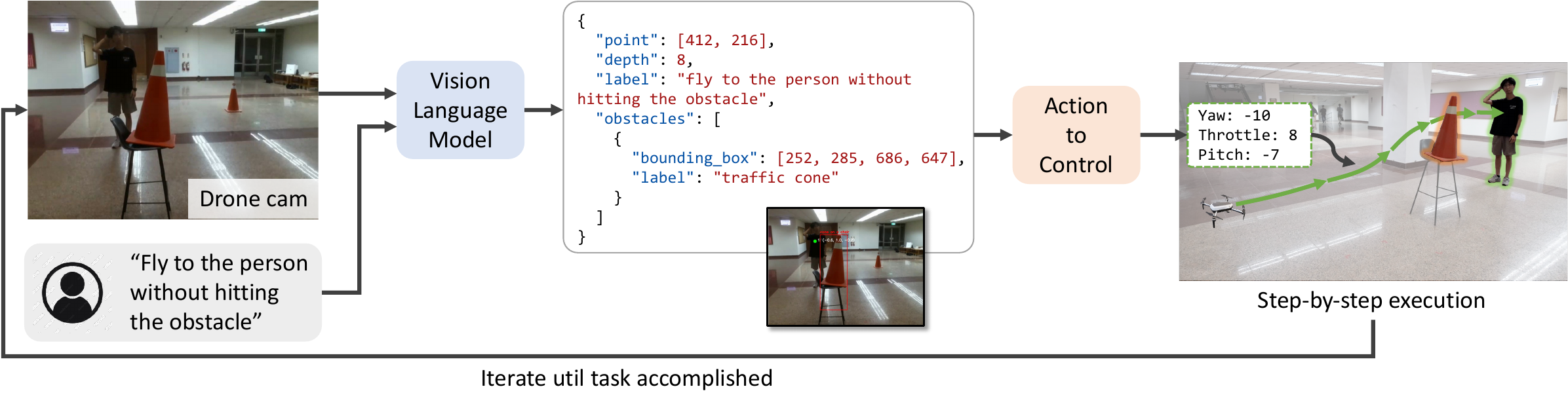}
    \vspace{-7mm}
    \caption{\textbf{Pipeline overview.} A camera frame and user instructions enter a frozen vision-language model, which returns a structured JSON with a 2D waypoint and any obstacle boxes. An Action-to-Control layer converts this output into low-level velocity commands (yaw, throttle, pitch) that steer the UAV. The loop repeats until the task is completed.
    }
    \label{fig:pipeline}
\end{figure}


We formulate UAV navigation as an iterative target-reaching process in 3D space. At each timestep \( t \), the system processes the current visual observation \( I_t \in \mathbb{R}^{H\times W\times3} \) along with a natural language instruction \( \ell \) to determine the next motion. Formally, we define a policy \( \pi(\cdot \mid \ell, I_t) \) that maps the observation-instruction pair to a 3D motion command \( m_t \in \mathcal{A} \), where the action space \( \mathcal{A} \subseteq \mathbb{R}^3 \) represents feasible displacement vectors.

We leverage vision-language models (VLMs) to implement the policy \( \pi \), transforming complex and vague navigation nature 
 language instructions into sequences of interpretable waypoint decisions. This approach decomposes the navigation task into discrete spatial reasoning steps that can be efficiently converted into UAV control signals, while remaining robust to diverse environments and instruction types.

As illustrated in Fig.~\ref{fig:pipeline}, our system runs an iterative perception-action loop with three stages:
(1) Given \( \ell \) and \( I_t \), we use the VLM \( G \) to produce a structured spatial understanding, 2D waypoints and moving step sizes (Sec.~\ref{sec:vlm_planning}),
(2) We transform the predicted 2D waypoint and step size into a 3D displacement vector, yielding executable low-level commands \( m_t \) (Sec.~\ref{sec:adaptive_scaling} and Sec.~\ref{sec:policy_mapping}), and
(3) A lightweight reactive controller continuously updates the observation, replans using the VLM, and executes the resulting motion commands in a closed-loop manner (Sec.~\ref{sec:control_loop}).

By outsourcing high-level spatial reasoning to the VLM and employing a lightweight geometric controller, our method achieves robust zero-shot UAV navigation directly from language---without relying on skill libraries, external depth sensors, policy optimization, or model training.

\subsection{VLM-based Obstacle-Aware Action Planning}
\label{sec:vlm_planning}

We frame the first stage of our method as a structured visual grounding task, where a VLM \( G \) processes an egocentric UAV camera observation \( I_t \in \mathbb{R}^{H \times W \times 3} \) alongside a natural language instruction \( \ell \) specifying the desired UAV task. Conditioned on this input, the VLM outputs a probability distribution \( P_G(w \mid \ell, I_t) \) over candidate waypoint plans \( w \in \mathcal{W}\), where \( \mathcal{W} \) represents the discrete space of feasible spatial waypoint sequences. We define the intermediate spatial plan \( O_t \) as the most likely waypoint sequence under this distribution:
\begin{equation}
O_t = \arg\max_{w \in \mathcal{W}} P_G(w \mid \ell, I_t).
\label{eq:vlm_plan_generation}
\end{equation}
The output \( O_t = \{u, v, d_\text{VLM}\} \) specifies a 3D navigation target in image space, where \( (u, v) \) are pixel coordinates and \( d_\text{VLM} \in \{1, 2, \ldots, L\} \) is a discretized depth label. Importantly, \( d_\text{VLM} \) represents the VLM's prediction of intended travel distance along the UAV's forward direction (positive y-axis in body frame), rather than a sensored depth measurement.

When obstacle-avoidance mode is activated, the VLM is further constrained to generate waypoints that guide the UAV toward the goal while avoiding intersection with detected object bounding boxes, promoting safe navigation through cluttered environments. By formulating UAV control through this visual grounding approach, we transform complex spatial reasoning into a computationally efficient task that enables robust, zero-shot, obstacle-aware navigation without requiring iterative optimization or exhaustive low-level action sampling.

\begin{figure}[t]
    \centering
    \vspace{-6mm}
    \includegraphics[width=1.0\linewidth]{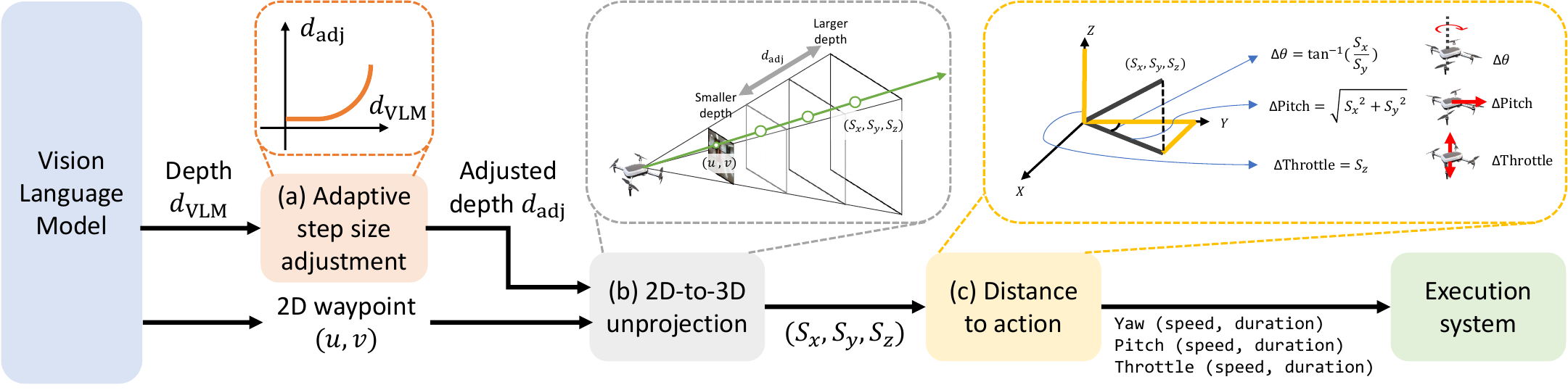}
    \vspace{-6mm}
    \caption{\textbf{Control-geometry details of our VLM-driven flight loop.}
    A frozen vision-language model first predicts a 2D waypoint $(u, v)$ and a discrete depth cue $d_\text{VLM}$. 
    (a) A nonlinear scaling curve converts $d_\text{VLM}$ into an adaptive step size $d_\text{adj}$, letting the UAV take larger strides in open space and smaller ones near obstacles.
    (b) The pair $(u, v, d_\text{adj})$ is unprojected through the pin-hole model to a 3D displacement vector $(S_x, S_y, S_z)$ in the UAV's body frame.
    (c) This vector is decomposed into control primitives: yaw $\Delta\theta=\text{tan}^{-1}(S_x/S_y)$, pitch $\Delta \text{Pitch}=\sqrt{{S_x}^2+{S_y}^2}$, and throttle $\Delta \text{Throttle}=S_z$. 
    These quantities are sent as timed velocity commands by the execution layer.
    The perception, planning, and control cycle repeats until the language instruction is fulfilled. 
    }
    \label{fig:method_details}
\end{figure}

\subsection{Adaptive Travel Distance Scaling}
\label{sec:adaptive_scaling}
Although VLMs can infer high-level spatial plans from visual inputs, they often lack a precise understanding of real-world 3D geometry and UAV navigation intuition possessed by human pilots. Consequently, motion commands derived directly from VLM outputs may result in overly aggressive or unsafe movements, particularly in cluttered or constrained environments.

To address this limitation, as shown in Fig.~\ref{fig:method_details} (a), we employ a non-linear scaling curve that converts the discrete depth label $d_\text{VLM}$ into an adjusted step size $d_\text{adj}$:
\begin{equation}
d_\text{adj} = \max \left( d_{\min},\; s \times \left( \frac{d_\text{VLM}}{L} \right)^p \right),
\end{equation}
where $s$ is a global scaling factor, $p$ controls the nonlinearity of the scaling curve, and $d_{\min}$ specifies a lower bound on the step size to ensure safety.  

This adaptive scaling approach enables the UAV to take larger steps in open areas while executing smaller, more cautious movements near targets and obstacles. The UAV can thus adapt its trajectory naturally to scene geometry without requiring explicit 3D maps or external depth sensors. This capability is particularly valuable for lightweight UAVs where onboard perception and strict latency constraints limit the feasibility of deploying traditional depth-sensing hardware.

\subsection{Policy Mapping from Image Space to 3D Actions}
\label{sec:policy_mapping}
Given the structured VLM output \( O_t = \{u, v, d_\text{adj}\} \), our system transforms this image-space waypoint into executable 3D motion commands. This transformation defines the core of our reactive policy, enabling the UAV to navigate toward visually grounded targets using only RGB inputs.

As depicted in Fig.~\ref{fig:method_details} (b), we unproject the predicted 2D waypoint $(u, v)$ together with the adjusted depth $d_\text{adj}$ through a pin-hole camera model to obtain a 3D displacement vector $(S_x, S_y, S_z)$, which is later decomposed into yaw, pitch and throttle commands.

To compute the desired 3D displacement vector $(S_x, S_y, S_z)$, the angular projection of the pixel location onto the camera's field of view is used:
\begin{equation}
S_x = u \cdot d_\text{adj} \cdot \tan(\alpha),\quad
S_y = d_\text{adj},\quad
S_z = v \cdot d_\text{adj} \cdot \tan(\beta),
\end{equation}
where \( \alpha \) and \( \beta \) are the camera's horizontal and vertical half field-of-view angles, respectively. The forward motion $S_y$ is aligned with the UAV's body-frame y-axis.

\subsection{Reactive Control Loop Execution}
\label{sec:control_loop} 
Operating within a closed-loop control framework, desired 3D displacements are decomposed into UAV control primitives: pitch, yaw, and throttle, as illustrated in Fig.~\ref{fig:method_details} (c). Each control primitive is converted into a velocity-duration pair, where the duration is derived from the magnitude of the required adjustment and a predefined constant speed. Commands are enqueued into an execution queue and sent to the UAV with temporal synchronization, allowing for smooth, responsive, and low-latency control through continuous correction. This approach enables efficient adaptation to dynamic environments without requiring complex trajectory optimization. For more technical details, please refer to the supplementary material.

%% file: 5_experiments.tex
\section{Experimental Results}
\label{sec:experiments}


\textbf{Experimental Setup.} We evaluated our approach in both simulated and real-world environments. For simulation, we employed the high-fidelity DRL simulator~\cite{DRLSim2024}, which serves as a standard benchmark from the Drone Racing League competition and effectively bridges the simulation-to-real gap through accurate physics modeling and realistic sensor simulation. For real-world validation, we implemented our system on a DJI Tello EDU drone platform, controlled through the Python SDK using low-level \texttt{rc} velocity commands. We conducted extensive tests across various indoor environments (office spaces, corridors, living areas) and outdoor settings (parks, campus walkways) with different lighting conditions, obstacle densities, and visual complexities to thoroughly assess real-world performance.

\textbf{Metrics.} We evaluated performance using two metrics: \textbf{Success Rate (SR)}, the percentage of trials where the drone reached its target without collisions, and \textbf{Completion Time}, measuring duration from movement initiation to task completion. These metrics together assess both reliability and efficiency across diverse navigation scenarios.



\textbf{Task Categories.} Our evaluation framework includes 6 distinct task categories designed to assess the robustness and versatility of VLM-guided UAV control across diverse navigation scenarios: (1) \textbf{Navigation:} Navigating to specified static targets or objects / locations in the real-world. (2) \textbf{Obstacle Avoidance:} Reaching designated targets while avoiding static and dynamic obstacles. (3) \textbf{Long Horizon:} Multi-stage navigation sequences requiring sustained performance and compositional planning across extended spatial and temporal scales. (4) \textbf{Reasoning:} Tasks requiring contextual interpretation, spatial inference, and environmental understanding beyond literal instruction following. (5) \textbf{Search:} Target localization tasks where targets initially lie outside the UAV's field of view. (6) \textbf{Follow:} Identifying and tracking real-world objects or people.

We design a total of 23 tasks for simulation and 11 tasks for real-world evaluation, across task categories. Each task was executed 5 times per method to account for execution variability. Performance metrics were aggregated by category to assess domain-specific capabilities. Complete task specifications and evaluation protocols are detailed in the supplementary material.




\textbf{Baselines.} We benchmark our approach against three representative methods for language-guided UAV control: (1) \textbf{TypeFly}~\cite{chen2023typefly}: A language-driven approach that uses GPT-4 to interpret natural language commands and select appropriate actions from a predefined skill library. While effective for known tasks, this method's reliance on a fixed action space fundamentally limits its zero-shot generalization capabilities; (2) \textbf{PIVOT}~\cite{nasiriany2024pivot}: A visual-language approach that overlays candidate 2D waypoints on the input image as visual prompts, from which a VLM selects the most appropriate waypoint for navigation. This approach requires pre-generating and evaluating multiple candidate paths rather than directly predicting optimal waypoints; (3) \textbf{Plain VLM}: An ablation of our method that directly prompts a VLM to predict drone actions in textual form without our proposed structured output formulation, spatial transformation, or adaptive depth scaling techniques.

We used the publicly released implementation for TypeFly, while PIVOT and Plain VLM were re-implemented following their published methodologies to ensure fair comparison within our evaluation framework.

\subsection{Performance Evaluation}
\label{sec:performance_eval}

\begin{table*}[t]
    \centering
    \vspace{-6mm}
    \caption{\textbf{Success rate (\%) comparison across task categories.} Our framework significantly outperforms TypeFly~\cite{chen2023typefly} and PIVOT~\cite{nasiriany2024pivot} baselines in both high-fidelity simulation and real-world DJI Tello experiments. We achieve 93.9\% and 92.7\% overall success rates in simulation and real-world settings, respectively. Note that Search tasks were exclusively evaluated in simulation, while Follow tasks were only tested in real-world settings due to environment constraints.}
    \label{tab:qualitative_results}
    \vspace{2mm}
    \resizebox{0.9\textwidth}{!}{
        \begin{tabular}{
            >{\raggedright\arraybackslash}p{0.17\linewidth} 
            >{\centering\arraybackslash}p{0.1\linewidth} 
            >{\centering\arraybackslash}p{0.1\linewidth} 
            >{\centering\arraybackslash}p{0.1\linewidth} 
            >{\centering\arraybackslash}p{0.1\linewidth} 
            >{\centering\arraybackslash}p{0.1\linewidth} 
            |>{\centering\arraybackslash}p{0.1\linewidth} 
        }
            \toprule
            \textbf{Method} & \textbf{Navigation} & \textbf{Obstacle Avoid} & \textbf{Long Horizon} & \textbf{Reasoning} & \textbf{Search / Follow} & \textbf{Overall Avg.} \\
            \midrule
            \multicolumn{7}{l}{\textit{Simulation}} \\ 
            \quad TypeFly \cite{chen2023typefly} & 1/25 & 0/25 & 0/25 & 0/15 & 0/25 & 0.9\% \\
            \quad PIVOT \cite{nasiriany2024pivot} & 11/25 & 4/25 & 7/25 & 2/15 & 9/25 & 28.7\% \\
            \quad \method{} (Ours) & \textbf{25/25} & \textbf{23/25} & \textbf{23/25} & \textbf{14/15} & \textbf{23/25} & \textbf{93.9\%} \\
            \midrule
            \multicolumn{7}{l}{\textit{Real-world}} \\
            \quad TypeFly \cite{chen2023typefly} & 1/5 & 3/10 & 5/10 & 2/20 & 2/10 & 23.6\% \\
            \quad PIVOT \cite{nasiriany2024pivot} & 0/5 & 1/10 & 0/10 & 2/20 & 0/10 & 5.5\% \\
            \quad \method{} (Ours) & \textbf{5/5} & \textbf{7/10} & \textbf{9/10} & \textbf{20/20} & \textbf{10/10} & \textbf{92.7\%} \\ 
            \bottomrule
        \end{tabular}
    }
    \vspace{-3mm}
\end{table*}

\begin{figure*}[t]
    \centering
    \resizebox{\textwidth}{!}{
        \begin{tabular}{ccc}
            \includegraphics[height=0.2\linewidth]{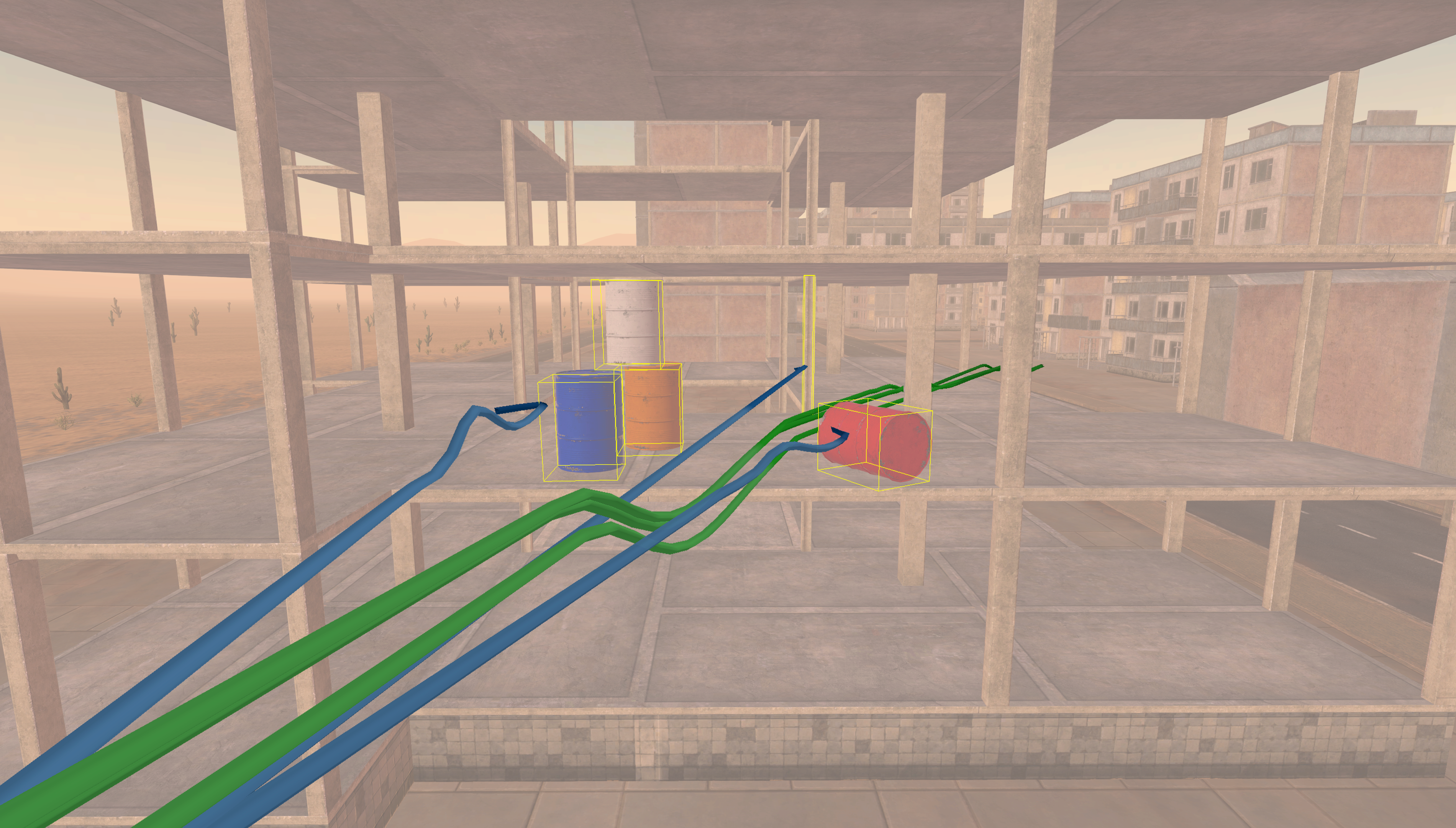} &
            \includegraphics[height=0.2\linewidth]{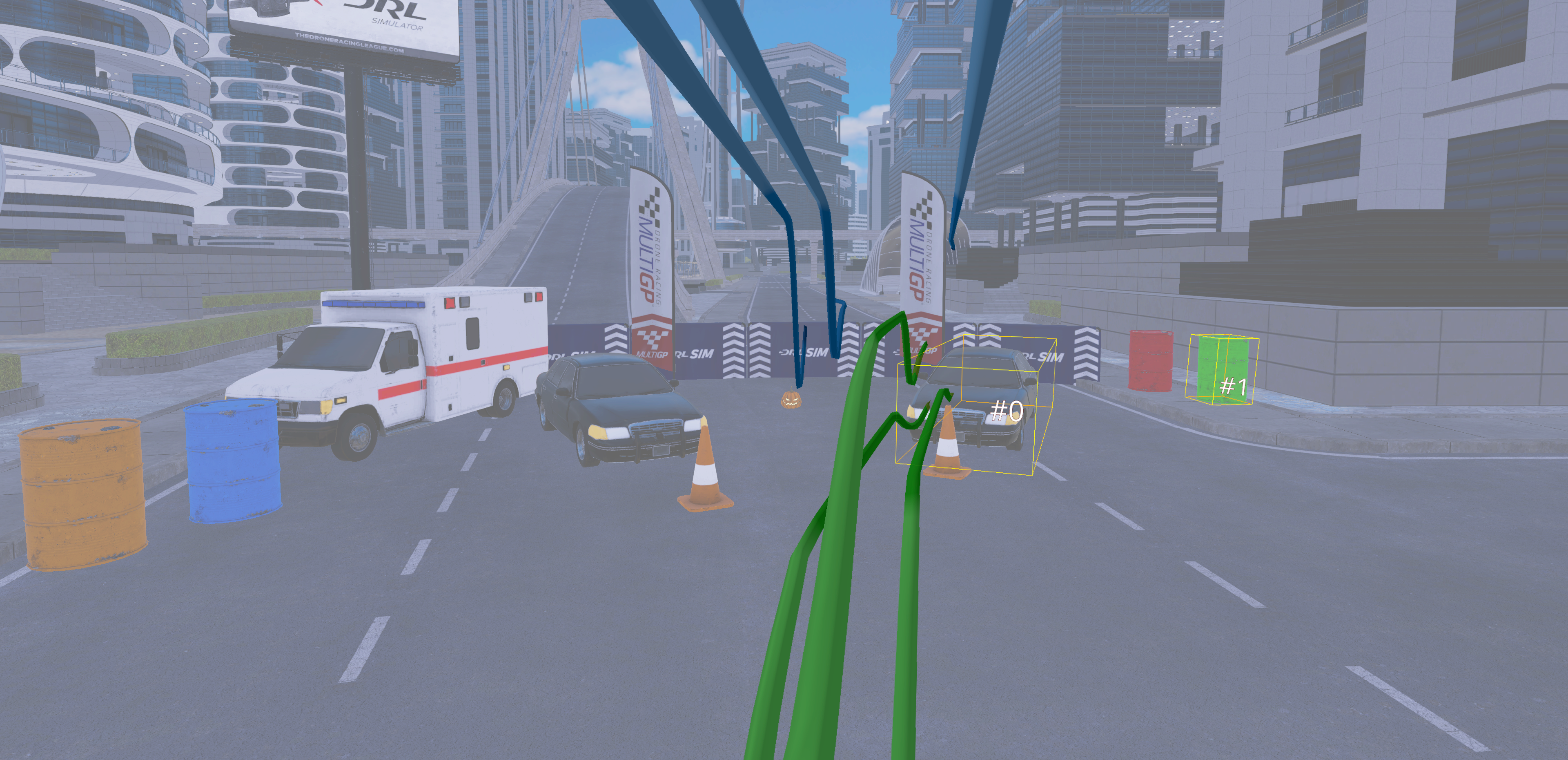} &
            \includegraphics[height=0.2\linewidth]{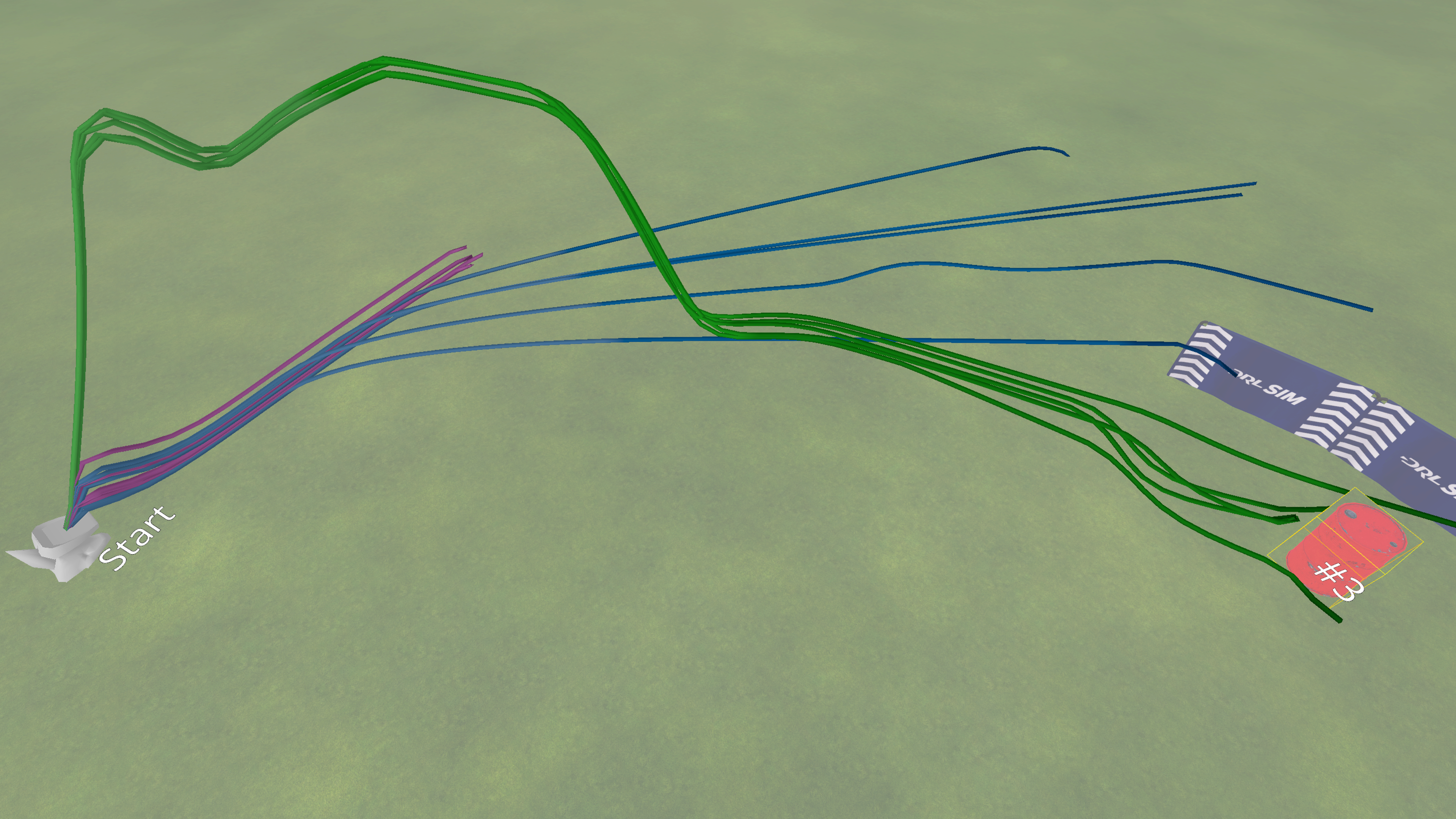} \\
            (a) Obstacle avoidance & (b) Target identification & (c) Pattern searching \\
        \end{tabular}
    }
    \vspace{-3mm}
    \caption{
        \textbf{Qualitative comparison of flight trajectories in the simulator.}
        Trajectory of our method is colored in \textcolor{teal}{\textbf{green}}, PIVOT \cite{nasiriany2024pivot} in \textcolor{blue}{\textbf{blue}}, and TypeFly \cite{chen2023typefly} in \textcolor{violet}{\textbf{purple}}. The absence of a colored path indicates the baseline failed to issue any fly command. Full videos are included in the supplementary materials.
    }
    \label{fig:qualitative_simulator}
    \vspace{-6mm}
\end{figure*}


We demonstrate the effectiveness of our method with the quantitative results shown in Table~\ref{tab:qualitative_results}. In simulation, our approach achieves \textbf{93.9\%} average success rate, significantly outperforming PIVOT (28.7\%) and TypeFly (0.9\%, limited by its predefined skill library). In particular, our framework excels in complex scenarios that require spatial reasoning and planning, such as obstacle avoidance (92\% vs. 16\% for PIVOT), long-horizon tasks (92\% vs. 28\% for PIVOT) and search tasks (92\% vs. 36\% for PIVOT).

\begin{figure*}[t]
\begin{minipage}{.69\textwidth}
    \setlength{\tabcolsep}{2pt}
    \small
    \resizebox{1.0\columnwidth}{!}{
        \begin{tabular}{ccccc}
            \raisebox{2.\normalbaselineskip}[0pt][0pt]{\rotatebox[origin=c]{90}{TypeFly}} & \includegraphics[width=0.25\columnwidth]{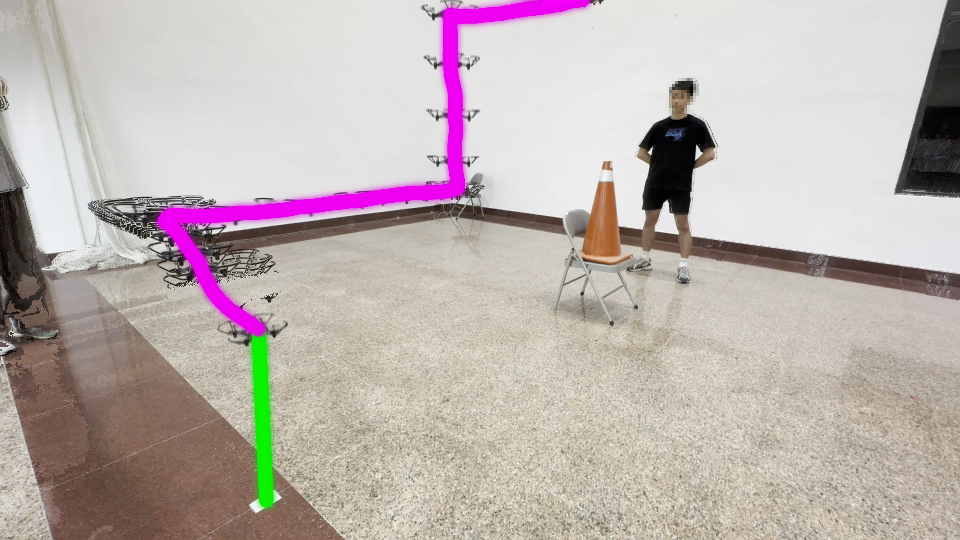} & \includegraphics[width=0.25\columnwidth]{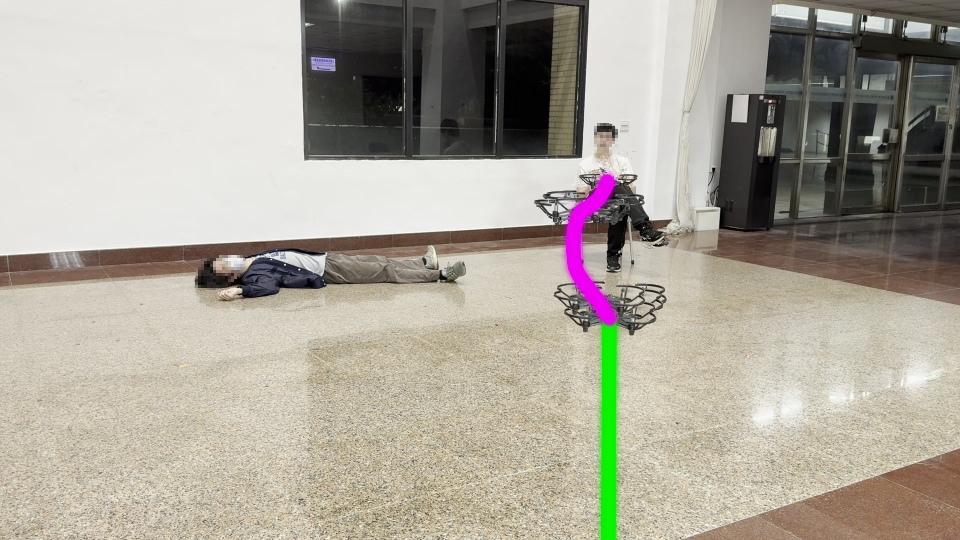} & \includegraphics[width=0.25\columnwidth]{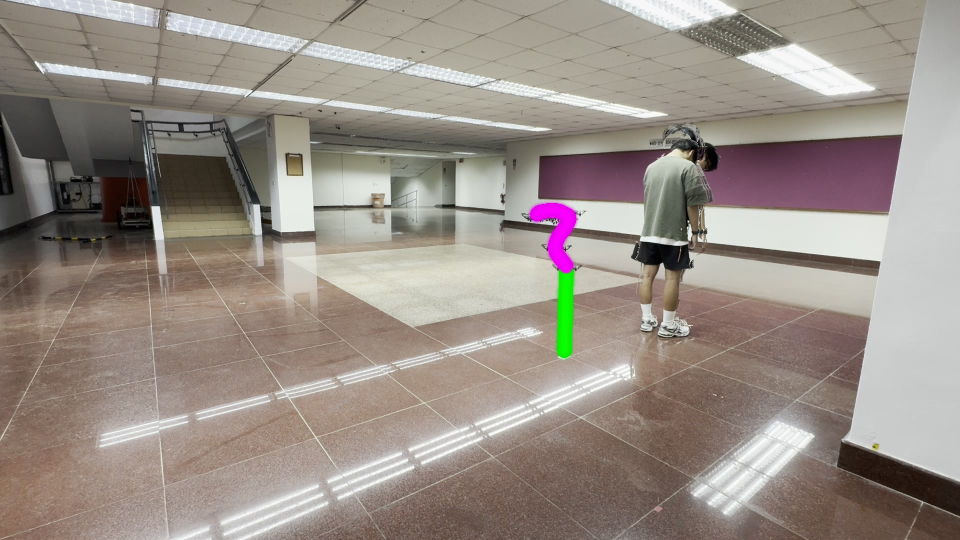} & \includegraphics[width=0.25\columnwidth]{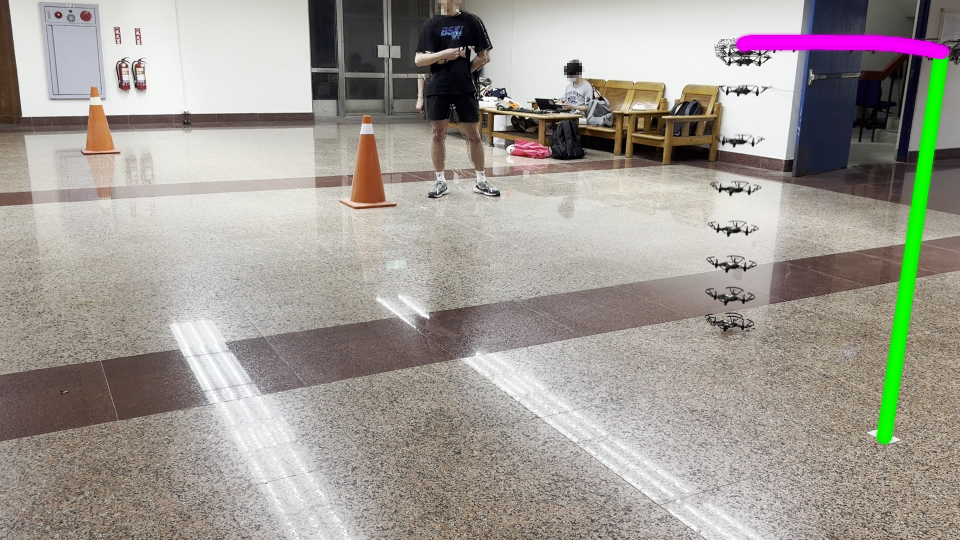} \\
            \raisebox{2.\normalbaselineskip}[0pt][0pt]{\rotatebox[origin=c]{90}{PIVOT}} & \includegraphics[width=0.25\columnwidth]{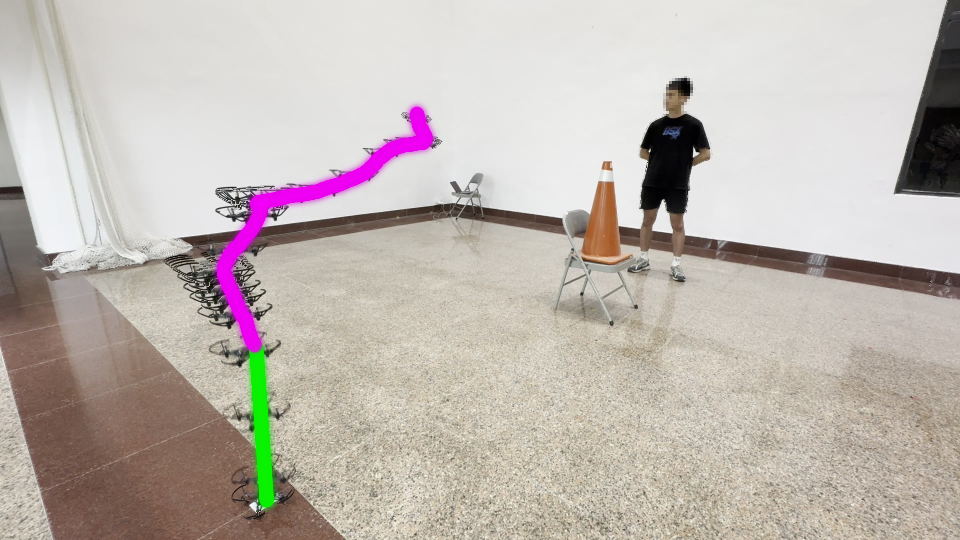} & \includegraphics[width=0.25\columnwidth]{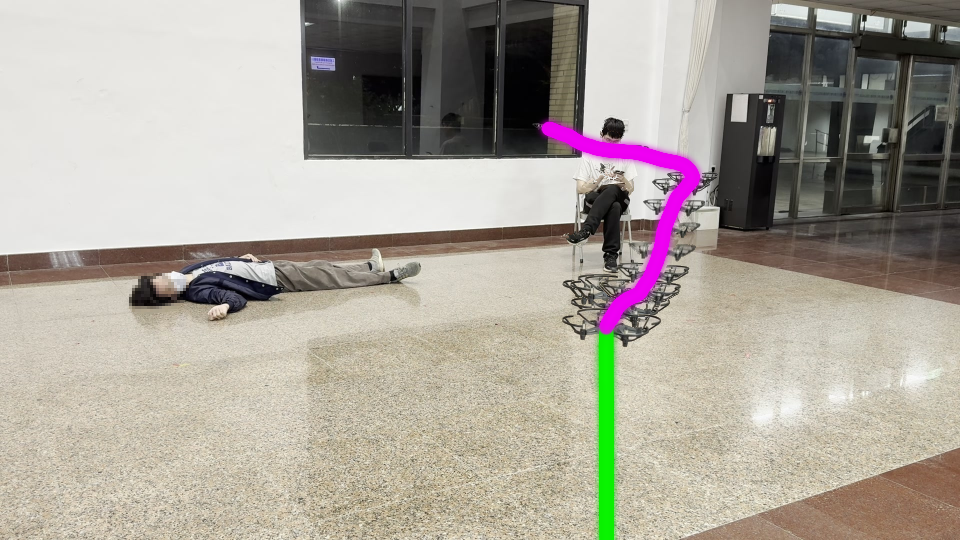} & \includegraphics[width=0.25\columnwidth]{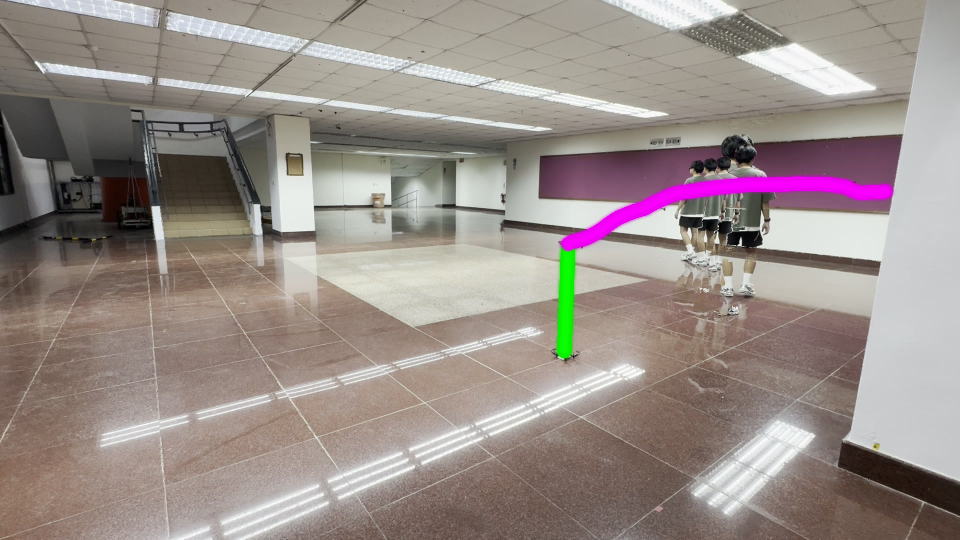} & \includegraphics[width=0.25\columnwidth]{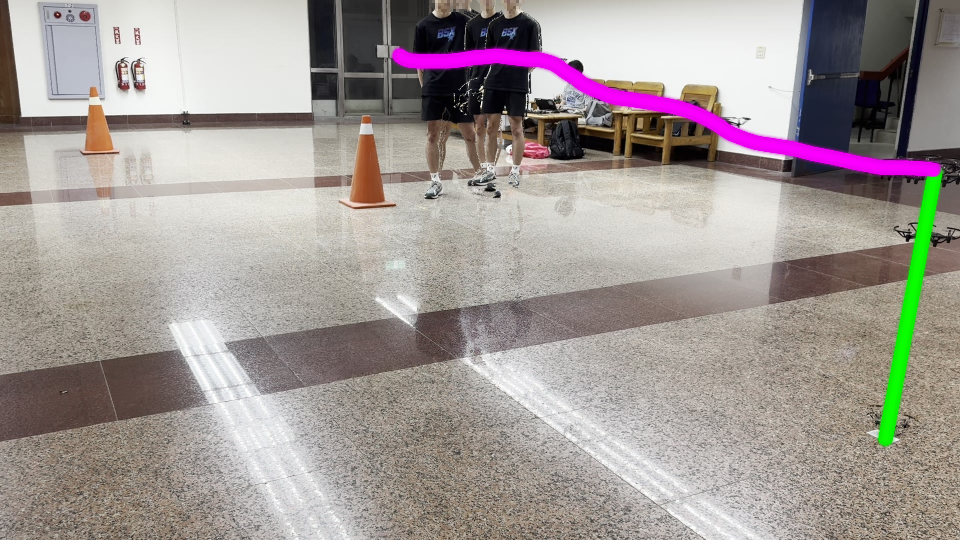} \\
            \raisebox{2.\normalbaselineskip}[0pt][0pt]{\rotatebox[origin=c]{90}{Ours}} & \includegraphics[width=0.25\columnwidth]{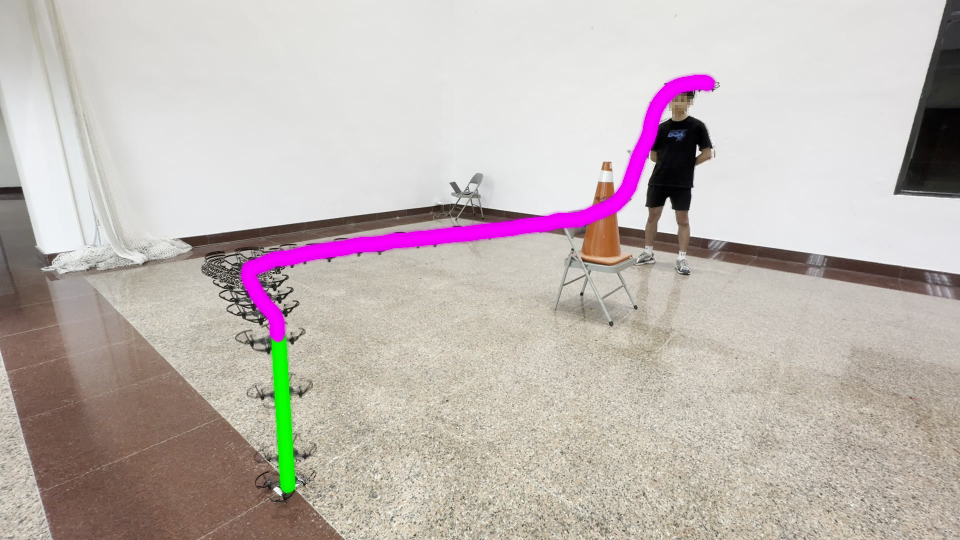} & \includegraphics[width=0.25\columnwidth]{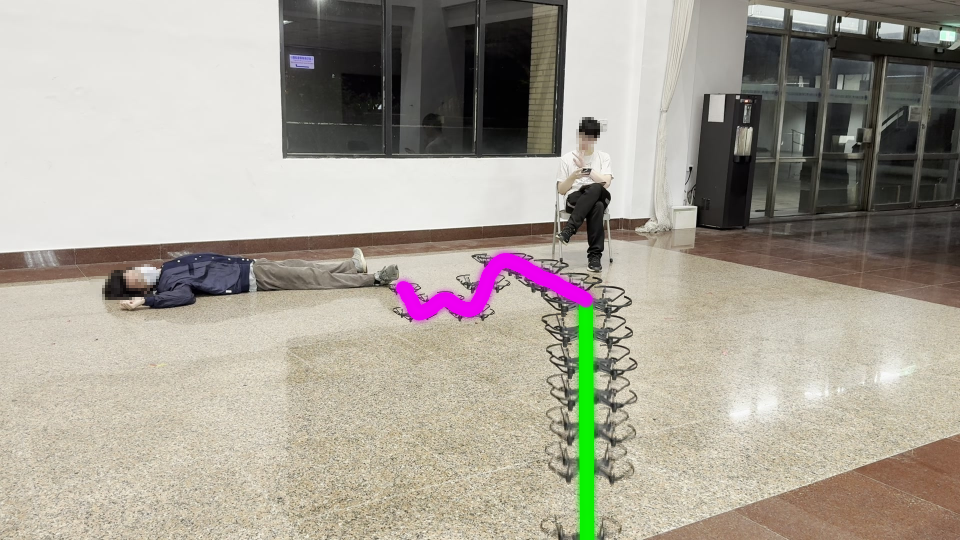} & \includegraphics[width=0.25\columnwidth]{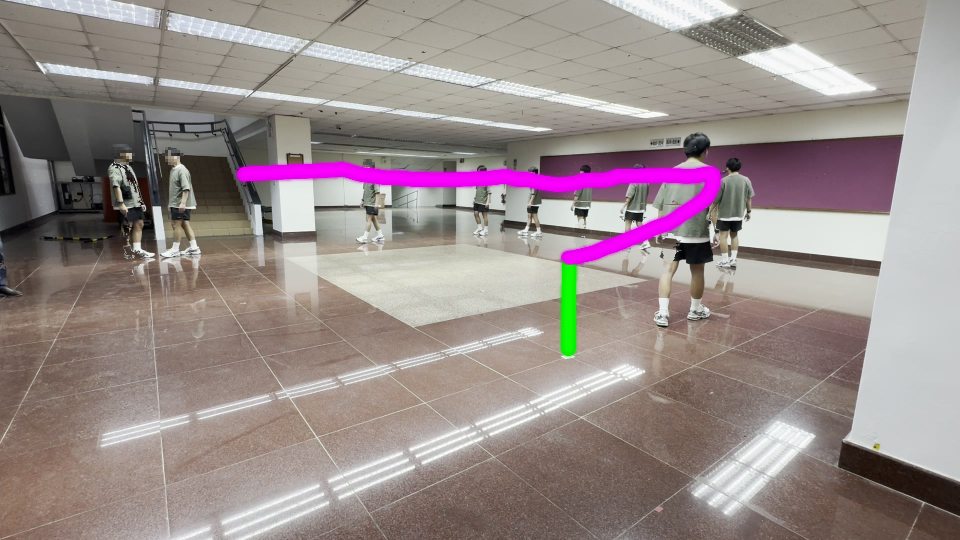} & \includegraphics[width=0.25\columnwidth]{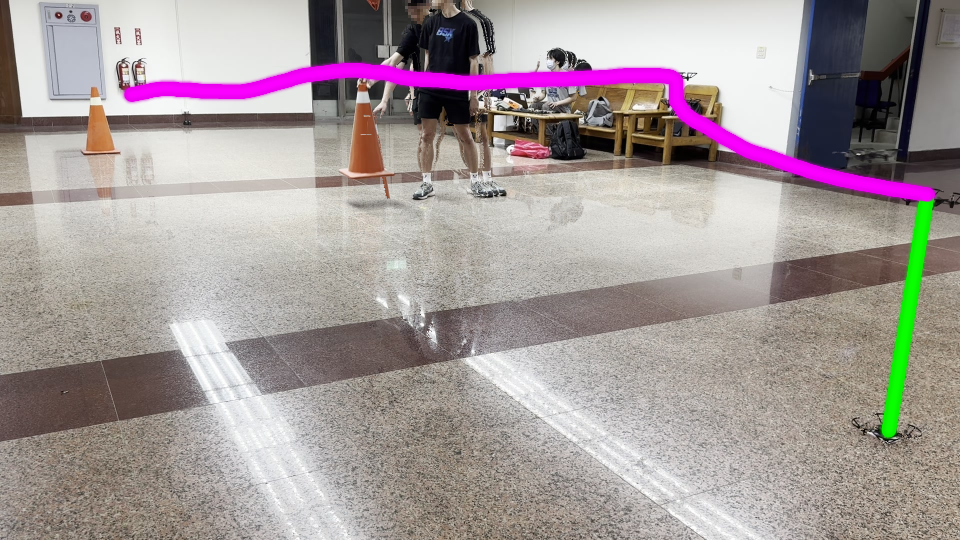} \\
        \end{tabular}
    }
    \vspace{-3mm}
    \caption{
        \textbf{Qualitative comparison of flight trajectories in the real-world.}
        Trajectory of our method compared to other baselines in the real-world testing. Take off trajectory is colored in \textcolor{green}{\textbf{green}} and task trajectory in \textcolor{magenta}{\textbf{magenta}}. Please refer to the supplementary materials for full videos.
    }
    \label{fig:qualitative_realworld}
\end{minipage}
\hspace{2mm}
\begin{minipage}{.28\textwidth}
    \includegraphics[width=\linewidth]{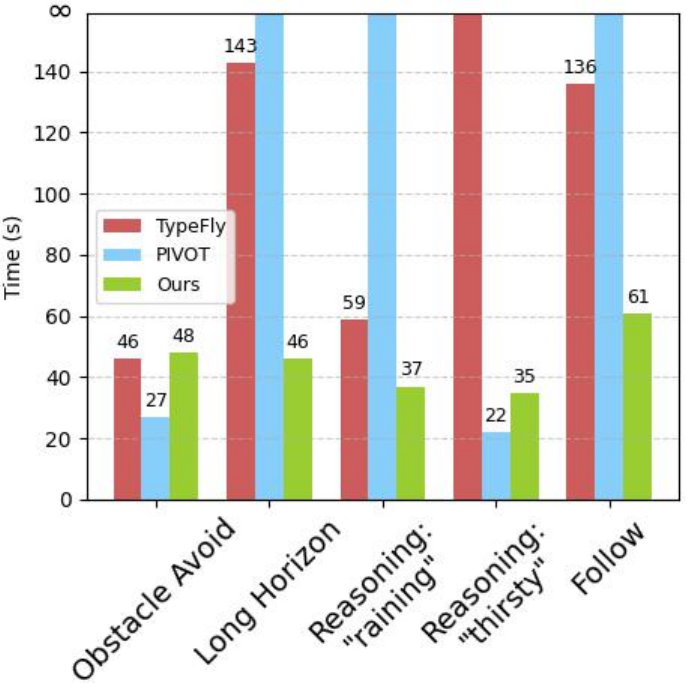}
    \vspace{-6mm}
    \caption{
        \textbf{Completion time by task.} Our approach achieves faster completion times across most tasks, particularly excelling in complex scenarios. Bars capped at $\infty$ indicate baseline failures.
    }
    \label{fig:qualitative_completion_time}
\end{minipage}
    \vspace{-3mm}
\end{figure*}

Real-world experiments confirmed our method's effectiveness with a \textbf{92.7\%} average success rate. In contrast, TypeFly struggled with object recognition and language understanding, while PIVOT performed poorly in real-world settings, demonstrating the advantages of our structured visual grounding approach.
We evaluated completion time across 5 representative real-world tasks including obstacle avoidance, long horizon, reasoning, and follow categories. As shown in Fig.~\ref{fig:qualitative_completion_time}, \method{} not only successfully completed all tasks where both baselines often failed, but also achieved faster completion times. These results demonstrate our method's superior efficiency and reliability in diverse scenarios.



We present qualitative results in simulation (Fig.~\ref{fig:qualitative_simulator}) and in the real-world (Fig.~\ref{fig:qualitative_realworld}). Our results suggest that our \method{} is more effective in generating smooth navigation trajectories, avoiding obstacles, and reaching the target than TypeFly and PIVOT.

\subsection{Ablations}
\label{sec:ablations}

\begin{table*}[t]
\begin{minipage}{.5\textwidth}
\centering
    \caption{
    \textbf{Ablation of prompting strategies and VLM backbones in simulation.} 2D$\to$3D waypoint prompting (\method{}) lifts SR from 40\% (PIVOT) to 87\% with Flash-Lite and hits 100\% on stronger VLMs, whereas plain text generation scores only 7\%.
    }
    \label{tab:ablation_summary}
    \vspace{2mm} 
    \resizebox{1.0\textwidth}{!} {
        \begin{tabular}{lcccc}
            \toprule
             & \textbf{Action prediction} & \textbf{VLM model} & \textbf{SR (\%)} \\
            \midrule
            Plain VLM & Text Generation & Gemini 2.0 Flash\cite{GoogleDeepMindGemini2024} & 7 \\
            PIVOT~\cite{nasiriany2024pivot} & Visual Prompting & Gemini 2.0 Flash\cite{GoogleDeepMindGemini2024} & 40 \\
            \midrule
            \multirow{6}{*}{\method{} (Ours)} & \multirow{6}{*}{2D Waypoint Labeling} & Gemini 2.0 Flash-Lite~\cite{GoogleDeepMindGemini2024}  & \textbf{87} \\
            & & Gemini 2.0 Flash~\cite{GoogleDeepMindGemini2024} & \textbf{100} \\
             & & Gemini 2.5 Pro~\cite{GoogleDeepMindGemini2024} & \textbf{100} \\
             & & GPT-4.1~\cite{OpenAIGPT4Turbo2024}& \textbf{100} \\
             & & Claude 3.7 Sonnet~\cite{AnthropicClaudeSonnet2024} & \textbf{93.3} \\
             & & Llama 4 Maverick~\cite{MetaLlama4_2024} & \textbf{93.3} \\
            \bottomrule
        \end{tabular}
    }
\end{minipage}
\hspace{2mm}
\begin{minipage}{.47\textwidth}
\centering
        \caption{
        \textbf{Adaptive step‐size controller cuts completion time while preserving success.} Across two representative tasks, switching from a fixed step to our adaptive scaling halves flight duration and raises the success ratio to 5/5.
        } 
        \label{tab:adaptive_step_ablation} 
        \vspace{2.5mm}
    \resizebox{1.0\textwidth}{!} {
        \begin{tabular}{
            lccc
        }  
            \toprule
            \textbf{Task} & \textbf{Step} & \textbf{Compl. time} & \textbf{SR} \\ 
            \midrule
            \multirow{2}{12em}{``Fly to the cones and the next.''}
             & Fixed & 61s & 5/5 \\
             & Adaptive & \textbf{28s} & \textbf{5/5} \\ 
            \midrule
            \multirow{2}{12em}{``I'm thirsty. Find something that can help me.''}
             & Fixed & 50.25s & 4/5 \\
             & Adaptive & \textbf{35.20s} & \textbf{5/5} \\ 
            \midrule
            \multirow{3}{12em}{``It’s raining. Head to the comfiest chair that will keep you dry.''}
             & Fixed & 47s & 5/5 \\
             & Adaptive & \textbf{30s} & \textbf{5/5} \\ 
             & \\
            \bottomrule
        \end{tabular}
    }
\end{minipage}
    \vspace{-3mm}
\end{table*}

     

We conducted an ablation study to evaluate the effectiveness of each model component in simulation. Our study includes five simulated tasks and three real-world tasks across different categories. The results are presented in Table~\ref{tab:ablation_summary}.

\textbf{Structured Prompting and Grounding.} We compared three VLM-based action prediction approaches: our method (prompting VLM to label 2D waypoints on images), plain VLM (predicting actions as text) and PIVOT (selecting from candidate 2D points on images). Our approach significantly outperforms alternatives with a success rate of 100\% versus just 7\% for plain VLM and 40\% for PIVOT on navigation tasks, demonstrating the effectiveness of our structured visual grounding formulation.


\textbf{VLM.} Our method performs robustly across multiple VLMs: Gemini 2.5 Pro, Gemini 2.0 Flash, and GPT-4.1 all achieved 100\% success rate; Claude 3.7 Sonnet and Llama 4 Maverick reached 93.3\%; and even Gemini 2.0 Flash-Lite achieved 87\%. This demonstrates our framework's effective generalization across vision-language models of varying capabilities.


\textbf{Adaptive Travel Distance Scaling.} Our method significantly speeds up the travel time using the proposed adaptive distance scaling. It maintains navigation performance, while reducing the average completion time from 50.25 to 35.20 seconds. The results are presented in Table~\ref{tab:adaptive_step_ablation}. We refer to the supplementary material for more details of the experimental setup.

        

\textbf{Our VLM-Integrated Approach.} Our approach generates bounding boxes directly from the Vision-Language Model (VLM) in a single pass, enabling zero-shot generalization and low latency. This offers critical advantages over specialized detectors limited by fixed vocabularies.

\begin{table}[h!]
\centering
\caption{\textbf{Design trade-off for obstacle avoidance.}}
\label{tab:compact_tradeoff_note}
\scriptsize 
\begin{tabular}{lccc}
\toprule
\textbf{Method} & \textbf{Latency} & \textbf{Accuracy (\%)} & \textbf{Generalization} \\
\midrule
Ours (VLM-integrated) & 1.077s & 88.8 & Zero-shot (any object) \\
+ External Detector (YOLOv8n)~\cite{yolov8_ultralytics} & 1.726s & 72.2 & Limited to known classes \\
\bottomrule
\end{tabular}

\end{table}

%% file: 6_conclusion.tex
\paragraph{Conclusion}
\label{sec:conclusion}
We presented \method{}, a training-free framework that repurposes frozen vision-language models for universal UAV navigation. By casting action prediction as 2D waypoint grounding, then geometrically lifting these points to 3D displacements, our method sidesteps task-specific data collection and policy optimization. A lightweight adaptive controller closes the perception-action loop, yielding smooth flights despite second-level VLM latency. Across 23 simulated and 11 real-world tasks, \method{} achieved \textbf{93.9\%} and \textbf{92.7\%} success rates, respectively, substantially outperforming TypeFly and PIVOT while remaining model-agnostic and hardware-friendly.

%% file: appendix.tex
\section{Overview}
\label{sec:overview}
In this supplementary material, we present additional details and results to complement the main manuscript, ``See, Point, Fly: A Learning-Free VLM Framework for Universal Unmanned Aerial Navigation'' (hereafter referred to as the main submission or the main paper).

Section~\ref{sec:method_details_new} elaborates on our proposed \method{} (\methodfull{}) framework, as detailed in the main submission. Section~\ref{sec:task_specs_new} outlines task specifications, evaluation protocols, and provides example prompts (Table~\ref{tab:prompt_examples_numbered}) for simulated and real-world tasks. Section~\ref{sec:implementation_details_new} details key implementation parameters, including adaptive scaling, VLM backend, control architecture, and latency. Section~\ref{sec:qualitative_videos_new} directs the reader to demonstration videos of qualitative results. Section~\ref{sec:ablation_study_new} presents an ablation study on our adaptive step-size controller, including setup, quantitative results (Table~\ref{tab:ablation_results}), and visual examples (Figure~\ref{fig:task_images}).

This supplementary material aims to provide a deeper understanding of our methodology (presented in the main paper), experimental rigor, and key component benefits. Our video viewer webpage attached along with this supplementary material also provides a comprehensive task demo for each type of real-world and simulator tasks.
\vspace{1em} 

\section{Method Details}
\label{sec:method_details_new} 
\subsection{Reactive Control Loop Execution}

\textbf{Input Variables:} Given the structured VLM output $O_t = \{U, V, d_{\text{adj}}\}$, as described in the main paper. We transfer $(U, V)$ into normalized value $(U_{\text{norm}}, V_{\text{norm}})$.

\medskip 
\textbf{Position Calculations:} We use $U_{\text{norm}}$ and $V_{\text{norm}}$ to calculate the desired 3D displacement vector $(S_x, S_y, S_z)$. Where $\alpha$ and $\beta$ are the camera's horizontal and vertical half field-of-view angles, respectively. These calculations follow the pinhole camera model detailed in the main submission (see Section 3.3, Eq. 2 of the main paper for the specific formulation used).
\[ S_x = U_{\text{norm}}\cdot d_{adj}\cdot \tan(\alpha), \quad S_y = d_{\text{adj}}, \quad S_z = V_{\text{norm}}\cdot d_{adj}\cdot \tan(\beta) \]

\medskip 
\textbf{Control Parameters:} We use the 3D displacement vector $(S_x, S_y, S_z)$ to calculate the UAV control primitive displacement: $(\Delta\theta, \Delta\text{Pitch}, \Delta\text{Throttle})$. The derivation of these control primitives from the 3D displacement vector is detailed in the main submission (see Section 3.4 and Figure 3c of the main paper for the specific formulation used).
\[ \Delta \theta = \tan^{-1}\left(\frac{S_x}{S_y}\right), \quad \Delta \text{Pitch} = \sqrt{S_x^2 + S_y^2}, \quad \Delta \text{Throttle} = S_z \]

\medskip 
\textbf{Duration Calculations:} Afterwards, we use $(\Delta\theta, \Delta\text{Pitch}, \Delta\text{Throttle})$ and Pre-defined speed $(P_{\text{yaw}}, P_{\text{pitch}}, P_{\text{throttle}})$ to calculate the Duration of each primitive $(D_{\text{yaw}}, D_{\text{pitch}}, D_{\text{throttle}})$.
\[ D_{\text{yaw}} = \frac{\Delta \theta}{P_{\text{yaw}}}, \quad D_{\text{pitch}} = \frac{\Delta \text{Pitch}}{P_{\text{pitch}}}, \quad D_{\text{throttle}} = \frac{\Delta \text{Throttle}}{P_{\text{throttle}}} \]
\textbf{Duration formula:} \[ \text{Duration} = \frac{\Delta \text{Distance}}{\text{Predefined\_Speed}} \]

\medskip 
\subsection{UAV Command Queue Implementation} Finally, we send enqueued \texttt{rc} commands to the UAV using a Python SDK (specifically DJITelloPy \citep{DJITelloPyLatest_misc}).
\begin{lstlisting}[language=Python, caption={Python-like pseudocode for UAV command queue}, label={lst:uav_commands}]
# send_rc_control(roll, pitch, throttle, yaw_rate)

# Yaw control
def yaw(Pyaw, Dyaw):
    action_queue.append(send_rc_control(0, 0, 0, Pyaw))
    time_sleep(Dyaw)
    action_queue.append(send_rc_control(0, 0, 0, 0)) # Stop yaw rate

# Pitch control
def pitch(Ppitch, Dpitch):
    action_queue.append(send_rc_control(0, Ppitch, 0, 0))
    time_sleep(Dpitch)
    action_queue.append(send_rc_control(0, 0, 0, 0)) # Stop pitch rate / reset pitch

# Throttle control
def throttle(Pthrottle, Dthrottle):
    action_queue.append(send_rc_control(0, 0, Pthrottle, 0))
    time_sleep(Dthrottle)
    action_queue.append(send_rc_control(0, 0, 0, 0)) # Reset throttle
\end{lstlisting}

\bigskip 

\subsection*{Annotations}
\begin{enumerate}
    \item $(U, V)$: 2D waypoint
    \item $(d_{\text{adj}})$: Adaptive step size
    \item $(U_{\text{norm}}, V_{\text{norm}})$: Normalized 2D waypoint
    \item $(S_x, S_y, S_z)$: 3D displacement vector
    \item $(\alpha, \beta)$: camera's (horizontal FOV, vertical FOV), where FOV is field-of-view angle
    \item $(\Delta\theta, \Delta\text{Pitch}, \Delta\text{Throttle})$: UAV control primitive displacements
    \item $(P_{\text{yaw}}, P_{\text{pitch}}, P_{\text{throttle}})$: Pre-defined speed of each primitive
    \item $(D_{\text{yaw}}, D_{\text{pitch}}, D_{\text{throttle}})$: Duration of each primitive
\end{enumerate}

\section{Experimental Details}
\label{sec:experimental_details_new}

\subsection{Task Specifications and Evaluation Protocols}
\label{sec:task_specs_new}
Task outcomes are classified as Success or Failure. A trial is a Failure if the UAV collides or if, at the task's completion, the target is not visible within the drone's egocentric camera view. A trial is a Success if, without collision, it has completed the specific task as the prompt requested (e.g., fly through the building), or if the target is clearly visible in the final egocentric view and the drone is positioned within 1 meter (real-world) or 1-5 meters (simulator) of the target. These criteria are consistent with common evaluation practices in the AVLN benchmarks (e.g. \citep{liu2023aerialvln, lee2024citynav, wang2024towards, gao2025openfly}).

A comprehensive table of examples of prompts (Table~\ref{tab:prompt_examples_numbered}) is provided to illustrate the detailed instructions for simulated and real-world tasks.

\begin{table*}[ht]
    \centering
    \caption{Numbered Example Prompts Used Across Task Categories. Each prompt is individually numbered and grouped by environment and category.}
    \label{tab:prompt_examples_numbered}
    \renewcommand{\arraystretch}{1.3}
    \resizebox{\textwidth}{!}{
    \begin{tabular}{
        >{\raggedright\arraybackslash}p{0.12\linewidth} 
        @{\hspace{2em}}
        >{\raggedright\arraybackslash}p{0.20\linewidth} 
        |>{\raggedright\arraybackslash}p{0.6\linewidth} 
    }
    \toprule
    \textbf{Environment} & \textbf{Category} & \textbf{Prompt} \\
    \midrule
    \multirow{25}{*}{Simulation}
        & \multirow{5}{*}{Navigation}
            & 1. Take off and fly to the red crane \\
        &   & 2. Take off and fly to the tall white building \\
        &   & 3. Take off and fly to the white needle \\
        &   & 4. Take off and fly to the black car \\
        &   & 5. Fly through the tunnel in front of you \\
    \cmidrule{2-3}
        & \multirow{5}{*}{Obstacle Avoidance}
            & 1. Take off and fly to the white needle (with obstacles) \\
        &   & 2. Take off and fly to the black car (avoiding obstacles) \\
        &   & 3. Fly through the tunnel in front of you \\
        &   & 4. Take off and navigate fly through the hollow building without hitting it \\
        &   & 5. Navigate through complex bridge structure \\
    \cmidrule{2-3}
        & \multirow{5}{*}{Long Horizon}
            & 1. Fly through the first gate and the second \\
        &   & 2. Fly to the first tower and then fly to the second tower \\
        &   & 3. Go around the tree first and then fly up the hill \\
        &   & 4. Look around the plane in front you and then fly to the crane \\
        &   & 5. Fly over the building in front of you and search the environment behind it \\
    \cmidrule{2-3}
        & \multirow{3}{*}{Reasoning}
            & 1. Take off and scan this city area \\
        &   & 2. Fly to an object that can be drive by people \\
        &   & 3. Fly to the train cart after the locomotion \\
    \cmidrule{2-3}
        & \multirow{5}{*}{Search / Follow}
            & 1. Take off and search for the monorail train \\
        &   & 2. Search for the red balloon and fly through each other \\
        &   & 3. Take off and search for the train in sight if not look around and find it \\
        &   & 4. Take off and search for the tower \\
        &   & 5. Take off and search for the lake if you cannot find it in sight look around and search for it \\
    \midrule
    \multirow{13}{*}{Real-world}
        & \multirow{1}{*}{Navigation}
            & 1. Fly to the chair (long distance) \\
    \cmidrule{2-3}
        & \multirow{2}{*}{Obstacle Avoidance}
            & 1. Fly to the person without hitting the cone \\
        &   & 2. Fly to the person without hitting the door \\
    \cmidrule{2-3}
        & \multirow{2}{*}{Long Horizon}
            & 1. Fly to the chairs and the next \\
        &   & 2. Fly to the cone and the next \\
    \cmidrule{2-3}
        & \multirow{4}{*}{Reasoning}
            & 1. It's raining, head to the comfiest chair that looks like it'll keep you dry! \\
        &   & 2. Fly to the person who needs help \\
        &   & 3. I'm thirsty, find something that can help me. \\
        &   & 4. Fly to the person in the dark area \\
    \cmidrule{2-3}
        & \multirow{2}{*}{Search / Follow}
            & 1. Fly toward the body of the person with red cone \\
        &   & 2. Fly toward the person with green shirt \\
    \bottomrule
    \end{tabular}
    }
\end{table*}

\subsection{Implementation Details}
\label{sec:implementation_details_new} 
Our system uses an adaptive scaling mechanism (detailed in the main paper, Section 3.2) with parameters $s=10$, $L=10$, $d_{\text{min}}=0.1$m and $p=1.8$. The control architecture operates asynchronously with VLM inference at $\approx 0.3 \sim 1$ Hz and low-level commands at $\approx 10$ Hz, resulting in an end-to-end latency of $\approx 1.5 \sim 3$ seconds, primarily due to VLM inference time. Unless otherwise specified, all experiments use Gemini 2.0 Flash \citep{GoogleDeepMindGemini2024} as the VLM backend.

\subsection{Qualitative Videos}
\label{sec:qualitative_videos_new}
The qualitative results of our experiments are provided as demonstration videos in the \texttt{exp\_results} directory of the supplementary materials. For convenient viewing, a video viewer webpage is included in the same folder. By opening \texttt{index.html} in any modern web browser, the videos corresponding to the various tasks and scenarios discussed in the paper can be easily browsed and viewed. This interface is intended to facilitate the evaluation of our approach and visually support the findings presented.

\subsection{Experiment Setup of Adaptive Travel Distance Scaling}
\label{sec:ablation_study_new} 

\begin{figure*}[htbp]
    \centering
    \begin{subfigure}[b]{0.3\textwidth}
        \centering
        \includegraphics[width=\linewidth]{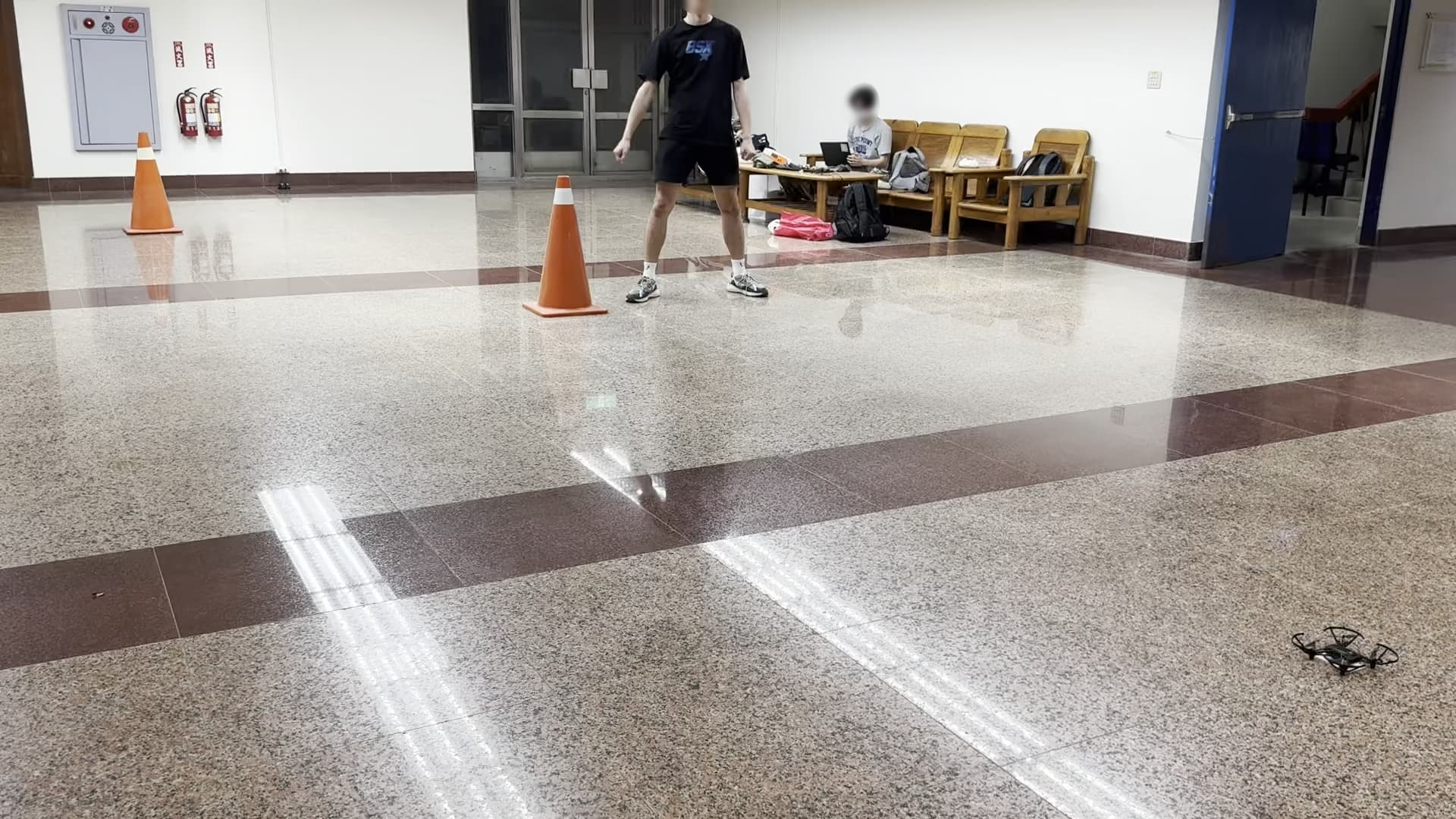}
        \vspace{1mm}
        Long Horizon
    \end{subfigure}
    \hfill
    \begin{subfigure}[b]{0.3\textwidth}
        \centering
        \includegraphics[width=\linewidth]{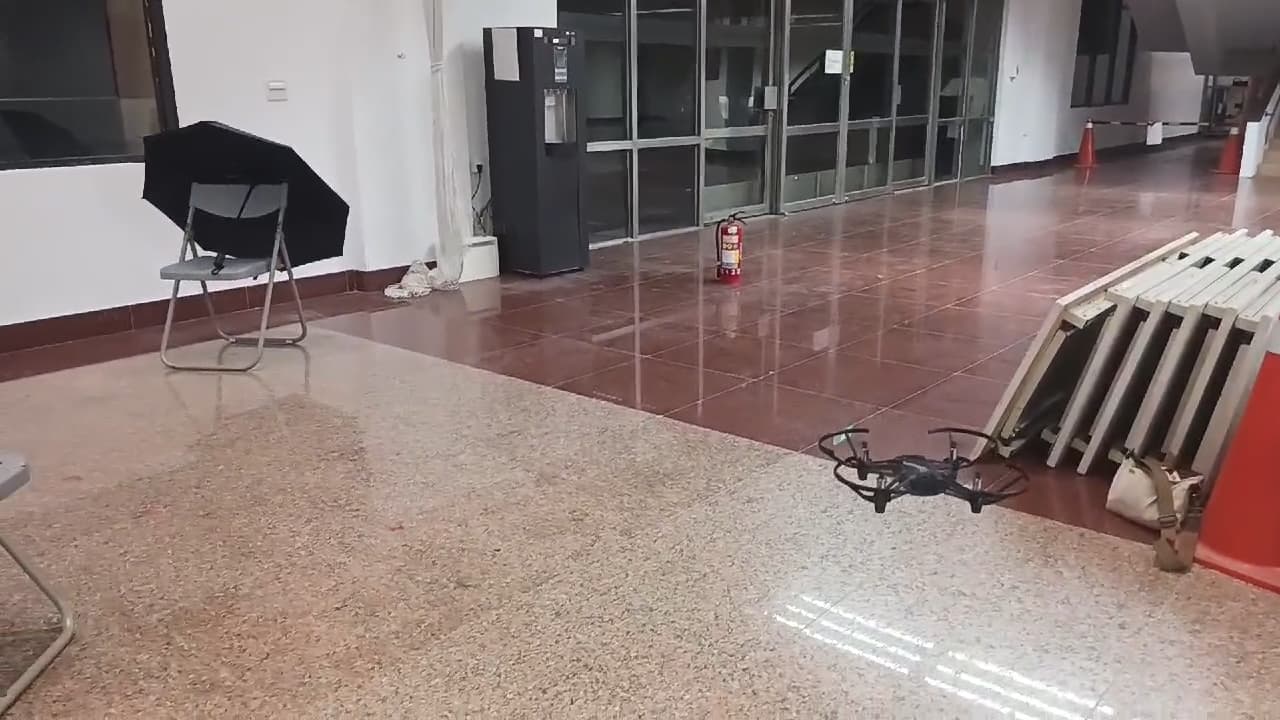}
        \vspace{1mm}
        Reasoning I
    \end{subfigure}
    \hfill
    \begin{subfigure}[b]{0.3\textwidth}
        \centering
        \includegraphics[width=\linewidth]{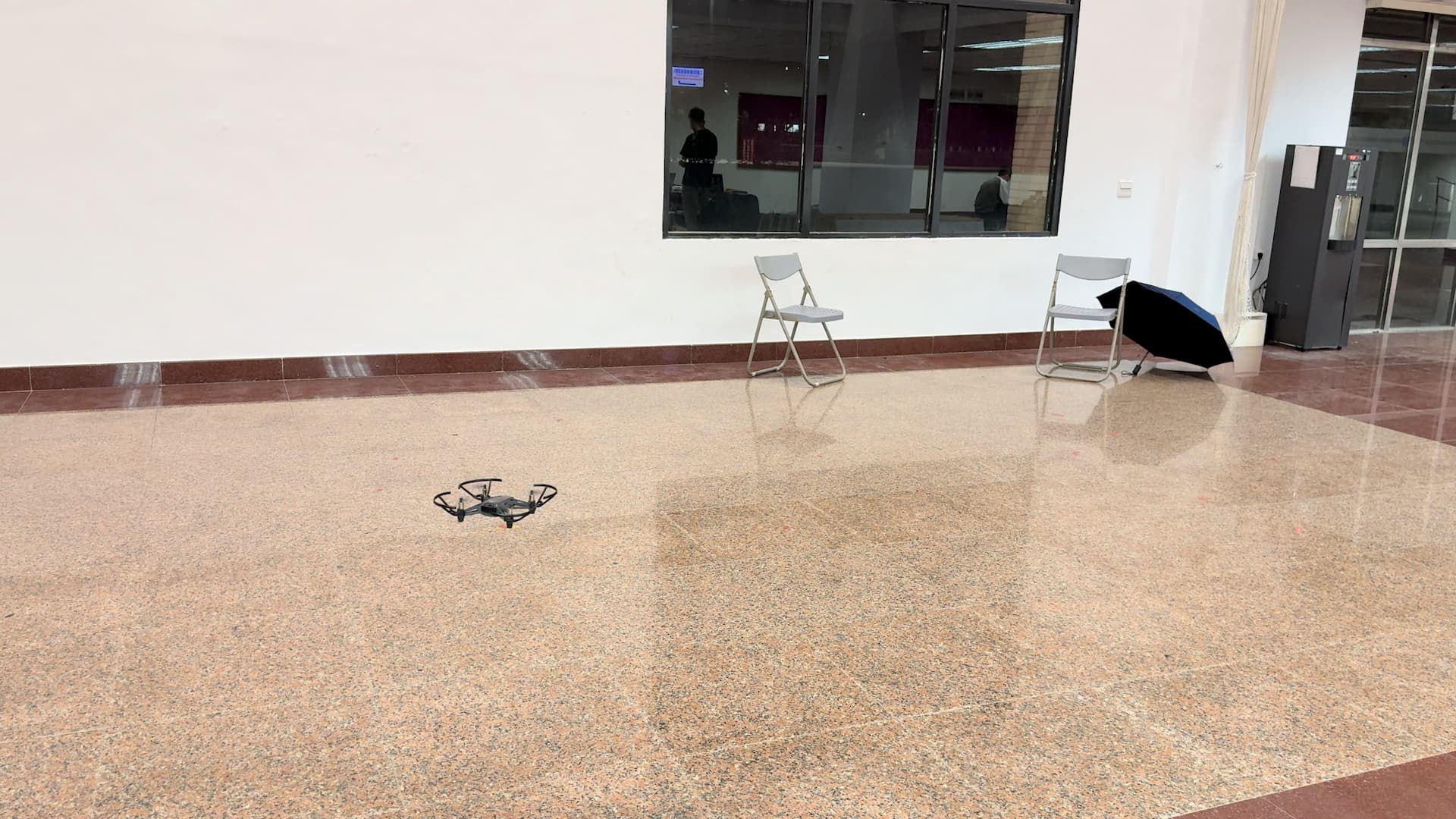}
        \vspace{1mm}
        Reasoning II
    \end{subfigure}
    \caption{Visual examples of the real-world scenarios for the tasks (from left to right): Long Horizon (``Fly to the cones and the next.''), Reasoning I (``I'm thirsty. Find something that can help me.''), and Reasoning II (``It's raining. Head to the comfiest chair that will keep you dry.''). These images depict the types of environments and objectives the UAV encountered during the ablation study evaluating the adaptive step-size controller.}
    \label{fig:task_images} 
\end{figure*}

To assess our adaptive step-size mechanism (Main Paper, Sec. 3.2), we conducted a real-world ablation study comparing our Adaptive Step-Size Controller against a fixed baseline. The experiments utilized a DJI Tello EDU drone, controlled via DJITelloPy \citep{DJITelloPyLatest_misc} using low-level \texttt{rc} velocity commands. Three distinct tasks, designed to test long-horizon planning and reasoning (detailed in Table~\ref{tab:ablation_results} and Figure~\ref{fig:task_images}), were each executed 5 times per controller configuration for robust comparison.


\begin{table}[H]
\centering 
\caption{Fixed vs. Adaptive Step-Size Controller performance on three real-world tasks. Metrics are Success Rate (SR) and Completion Time (Compl. time: start to finish). The adaptive controller significantly reduces completion times while maintaining or improving SR against the fixed baseline.}
\label{tab:ablation_results} 

\vspace{0.5em} 

\renewcommand{\arraystretch}{1.2}
\begin{tabular}{
    >{\raggedright\arraybackslash}p{0.52\linewidth} 
    @{\hspace{1.0em}}
    >{\raggedright\arraybackslash}p{0.16\linewidth} 
    @{\hspace{1.0em}}
    >{\centering\arraybackslash}p{0.13\linewidth} 
    >{\centering\arraybackslash}p{0.075\linewidth} 
}
\toprule
Prompt & Controller type & Compl. time & SR (\%) \\
\midrule
\multirow{2}{=}{Fly to the cones and the next.} & Fixed & 61.00s & 100 \\
& \textbf{Adaptive} & \textbf{28.00s} & \textbf{100} \\
\midrule
\multirow{2}{=}{I'm thirsty. Find something that can help me.} & Fixed & 50.25s & 80 \\
& \textbf{Adaptive} & \textbf{35.20s} & \textbf{100} \\
\midrule
\multirow{2}{=}{It's raining. Head to the comfiest chair that will keep you dry.} & Fixed & 47.00s & 100 \\
& \textbf{Adaptive} & \textbf{30.00s} & \textbf{100} \\
\bottomrule
\end{tabular}
\end{table}

The results in Table~\ref{tab:ablation_results} show that the adaptive controller significantly reduces task completion times while maintaining or improving success rates (SR). For instance, in the task ``Fly to the cones and the next.'' the completion time was more than halved (61s to 28s) with 100\% SR. For ``I'm thirsty. Find something that can help me.'', the adaptive controller decreased the completion time (50.25s to 35.20s) and improved the SR from 80\% to 100\%. The ``It's raining...'' task also saw a substantial time reduction (47s to 30s) with 100\% SR. These findings confirm the efficacy of the adaptive mechanism in enhancing operational efficiency and reliability in complex real-world settings.